\documentclass[journal]{IEEEtran} %[journal] instead of draft

\usepackage[utf8]{inputenc}         % allow utf-8 input
\usepackage[T1]{fontenc}            % use 8-bit T1 fonts
\usepackage{hyperref}               % hyperlinks
\usepackage{url}                    % simple URL typesetting (links)
\usepackage{nicefrac}               % compact symbols for 1/2, etc.
\usepackage{microtype}              % microtypography
\usepackage{amsfonts}               % blackboard math symbols
\usepackage{amsmath}                % For math
\usepackage[dvipsnames]{xcolor}     % For colored text
\usepackage{float}                  % For using H in images
\usepackage{graphicx}               % Specifies path and format of images
\graphicspath{{images/}}
\DeclareGraphicsExtensions{.pdf,.PDF,.jpg,.JPG,.jpeg,.JPEG,.png,.PNG}
\usepackage[caption=false]{subfig}  % Subfigure
\usepackage{booktabs}               % professional-quality tables
\usepackage{multicol}               % Figure on multicolumns 
\usepackage{multirow}               % Figure on multirow
\usepackage{booktabs}               % Fancy tables
\usepackage{array}                  % Fancy tables
\newcolumntype{C}[1]{>{\centering\arraybackslash}p{#1}}

\usepackage[nolist]{acronym}        % Abbreviations
\usepackage{verbatim}               % Comments
\usepackage{comment}                % Comment environment

\usepackage{textcomp}
\usepackage[numbers]{natbib}
\usepackage{makecell}

\usepackage{xcolor}

\newcommand\tocite[1]{\textcolor{blue}{[REFERENCE]}}

\begin{document}

% Paper title
\title{Identifying and Mitigating Flaws of Deep Perceptual Similarity Metrics}

%\begin{comment}
\author{
    \IEEEauthorblockN{
        \textbf{Oskar~Sjögren}\IEEEauthorrefmark{1}\IEEEauthorrefmark{2}, \and
        \textbf{Gustav~Grund~Pihlgren}\IEEEauthorrefmark{1}\IEEEauthorrefmark{2}, \and
        \textbf{Fredrik~Sandin}\IEEEauthorrefmark{1},
        \and \textbf{Marcus~Liwicki}\IEEEauthorrefmark{1}
    }\\
    \vspace{0.2cm}

    \IEEEauthorblockA{
        \IEEEauthorrefmark{1}%
        \textit{Machine Learning Group} \\
        Lule{\aa} University of Technology, Sweden\\
        %\{firstname\}.\{lastname\}@ltu.se\\
        \vspace{0.15cm}
        \IEEEauthorblockA{
        \IEEEauthorrefmark{2}
        \textit{Equal contribution}
    }\\
    }
}
%\end{comment}

\begin{comment}
\author{
    \IEEEauthorblockN{
        \textbf{Author A}\IEEEauthorrefmark{2}, \and
        \textbf{Author B}\IEEEauthorrefmark{2}, \and
        \textbf{Author C}\IEEEauthorrefmark{2}, \and
        \textbf{Author D}\IEEEauthorrefmark{2}, \and
        \textbf{Author E}\IEEEauthorrefmark{2}\IEEEauthorrefmark{4}
    }\\
    \vspace{0.2cm}
    \IEEEauthorblockA{
        \IEEEauthorrefmark{2}%
        \textit{affiliation} \\
        affiliation address \\
        affiliation emails \\
        \vspace{0.15cm}
        \IEEEauthorrefmark{4}%
        \textit{affiliation} \\
        affiliation address \\
        affiliation emails \\
    }
}
\end{comment}

% after the title page when using the two side option.
\markboth{}{First author et al. : Title}

% Make the title area
\maketitle

\thispagestyle{empty}

%%*************************************************************************
% Abstract
\begin{abstract}
% Typically you would want to have one sentence for the following concepts:
% - The problem tackled
% - Why nobody else has adequately answered the research question yet
% - How you tackled the research question
% - How did you go about doing the research that follows from your big idea
% - What’s the key impact of your research?

%The foundation of comparisons consists of similarity metrics, algorithms for calculating how similar two or more objects are.
Measuring the similarity of images is a fundamental problem to computer vision for which no universal solution exists.
%In the computer vision domain it is well known when and how historically popular metrics, such as the pixel-wise L2-norm, works.
While simple metrics such as the pixel-wise L2-norm have been shown to have significant flaws, they remain popular.
One group of recent state-of-the-art metrics that mitigates some of those flaws are Deep Perceptual Similarity (DPS) metrics, where the similarity is evaluated as the distance in the deep features of neural networks.
However, DPS metrics themselves have been less thoroughly examined for their benefits and, especially, their flaws.
This work investigates the most common DPS metric, where deep features are compared by spatial position, along with metrics comparing the averaged and sorted deep features.
%how they perform on known hard cases, and further when and why they perform poorly.
The metrics are analyzed in-depth to understand the strengths and weaknesses of the metrics by using images designed specifically to challenge them.
This work contributes with new insights into the flaws of DPS, and further suggests improvements to the metrics.
An implementation of this work is available online.\footnote{\url{https://github.com/guspih/deep_perceptual_similarity_analysis/}}
\end{abstract}

% Introduction 
\section{Introduction}
\label{toc:introduction}

%establish a term for the similarity metric to further be refered to as PS (perceptual simililarity) som gustav gjorde.

%This is an example of acronym usage for single \ac{AI} and multiple \acp{AI}. TODO: Remove

% Basically another abstract, but this time longer and with references.

% The problem tackled
% Why is it a problem on the first place i.e. why did not someone fix it already? In other words: what makes it hard?

% Why would one care about this problem?
% Why are you even making this ? Why does it makes sense? 

% Why are we performing the study?

% Explain how we limit the study.

% In the end, what do you bring new? Why would one care?

%Measuring the similarity of images is a fundamental problem to computer vision for which no universal solution exists.
%For example, whether a green triangle is more similar to a green circle or a red triangle is subjective and depend on the goal of the measurement.
%One such goal is to mimic the human perception of similarity and measurements that strives to achieve this are called perceptual similarity metrics.

Similarity metrics are a fundamental part of many machine learning processes.
Every time two or more objects are compared, a similarity metric is used.
In computer vision, widely used metrics, such as the pixel-wise L2-norm, have been carefully studied and their benefits and flaws are well-known which lets users make an informed decision when using them.

Many improvements to pixel-wise metrics have been, with a common goal being to mimic human perception with a so-called perceptual similarity.
One popular perceptual similarity metric is the Structural Similarity Index Measure~\cite{wang2004SSIM}.
A more recent approach is to utilize deep features learned by machine learning models for measuring perceptual similarity.
This practice, called Deep Perceptual Similarity (DPS) measures the similarity of two images by comparing their respective activations in the deep layers of Convolutional Neural Networks (CNNs), instead of using the pixel values directly.

DPS metrics have outperformed previous models on perceptual similarity~\cite{zhang2018unreasonable}.
Additionally, such metrics have been used as part of the loss function for training models, which have achieved impressive results on a host of tasks.
These tasks include, image generation~\cite{LarsenSW15}, style transfer and super-resolution~\cite{johnson2016perceptual}, object detection~\cite{li2017perceptual}, and image segmentation~\cite{mosinska2018beyond}.

While there are clear benefits of DPS, its flaws are not as well studied.
While it has been shown that deep perceptual similarity is vulnerable to adversarial examples, this is expected from any method depending on deep networks and existing methods for protecting from adversarial attacks such as ensembles may be utilized~\cite{Kettunen}.
Additionally, adversarial examples are quite complex compared to the known flaws of other metrics.
For example, pixel-wise metrics would consider a black-and-white image to be as dissimilar as possible from its inverted version.

This work aims to analyze if and how DPS can successfully handle the flaws of the pixel-wise L2-norm, and investigate if there are any similar unexplored flaws of DPS and how those may be mitigated.
Additionally, several different DPS metrics are analyzed for flaws and then evaluated on the BAPPS dataset~\cite{zhang2018unreasonable}, to check if those flaws translate into performance on an actual dataset.

The investigation of DPS is performed by creating image pairs that are similar to each other compared to some reference images and checking in which cases the DPS metrics succeed or fail in identifying the image pairs as more similar than the reference.
The feature maps of the CNNs used for calculating similarity are analyzed to gain insight as to what underlies the successes and failures.

\section{Related Work}
\label{toc:related_work}

%% Similarity metric development
Commonly used image similarity metrics are pixel-wise metrics where each pixel of one image is compared directly against the corresponding pixel of the other.
These metrics have long been known to be poor similarity metrics as they disregard high-level image structures~\cite{taylor1991measures,PWL-love-hate,zwang}.
%Instead perceptual similarity metrics have traditionally relied on measures such as brightness, contrast, 
Instead many different perceptual similarity metrics have been proposed including Dynamic Partial Function~\cite{li2003discovery}, the Structural Similarity Index Measure~\cite{wang2004SSIM}, and Structural Texture Similarity~\cite{zhao2008structural}.
Despite known flaws and suitable alternatives, per-pixel metrics have consistently been used for image comparison within computer vision in general, and to calculate the loss for machine learning models specifically.

%% Deep Perceptual Loss and Similarity
One powerful attribute of deep learning is that the deep features learned by the networks typically contain information useful for other tasks than the one the network was trained for.
This attribute was used to great effect with the introduction of neural style transfer, where the content and style of images were compared using different sets of deep features within a neural network~\cite{gatys2016image}.
This practice of training models to minimize the difference between the activations of a deep network in order to get visually similar images is known as deep perceptual loss.

Deep perceptual loss has since its introduction been successfully applied to a large number of computer vision tasks such as improving the performance of variational autoencoders~\cite{hou,Gustav,Bhard}, Generative Adversarial Networks~\cite{LarsenSW15}, Super-Resolution~\cite{dong,Lucas}, and style transfer~\cite{johnson2016perceptual}.
The method has been proven effective at the task of perceptual similarity where it significantly outperformed previous methods~\cite{zhang2018unreasonable}.
%The deep features of neural networks, as it turns out, are great as perceptual similarity measurement, even when the network is trained in a completely unsupervised fashion.
This method of calculating perceptual similarity using the deep features of neural networks is referred to as deep perceptual similarity (DPS).

%% Adversarial examples
One potential problem with DPS is that it relies on deep neural networks, which are known to be vulnerable to adversarial examples.
Adversarial examples are almost imperceptible perturbations to images or other input data that induce significant changes or errors to the prediction model~\cite{Ilyas2019}.
While no perfect protection from adversarial examples is currently known, there is a wide array of defenses that can be used, including using ensembles~\cite{Kettunen}.
Additionally, outside of malicious attacks, this is rarely a problem.

%% Other ways to improve similarity performance
% LPIPS/E-LPIPS
Another paradigm for creating similarity metrics is to optimize a machine learning model for the task~\cite{ricci1995learning}.
This has been applied to DPS with the LPIPS method, though it notably only performed marginally better than using methods that had only been pretrained~\cite{zhang2018unreasonable}.
Like with many other machine learning methods the results can be improved somewhat with the use of ensemble methods, though still comparable to pretrained models~\cite{Kettunen}.

% Google paper
Where this work analyzes DPS through deep analysis of cases where it fails, another recent work investigates how different network architectures and pretraining procedures affect performance~\cite{kumar2022surprising}.
That work found, among other things, that better pretraining performance on ImageNet~\cite{deng2009imagenet}, does not necessarily lead to better perceptual similarity,
It additionally showed that a good pretrained model can outperform models trained specifically for the similarity task.

% Feature map analysis
As DPS metrics inherently rely on the deep activations of neural networks, most commonly CNNs, analyzing these activations is inherently interesting.
Many methods for such analysis exist and one of the most common is to visualize the feature maps of the CNNs~\cite{zeiler2014visualizing}, which is utilized in this work.

\section{Deep Perceptual Similarity}
\label{toc:dps}

%Name suggestions for Beta-factor
%   Translation Invariant Perceptual Similarity (TIPS)
%   Sorted Deep Feature Similarity (SDFS) 
%   

% More details about everything to enable reproduction of our work
% Clarify that this subsection is mainly for those interested in the exact implementation or want to implement the work themselves

%By applying the L2 norm of the sorted features to the perceptual similarity metric, a translation invariant factor is added.
%The combination of perceptual loss and a sorted deep perceptual similarity factor makes the metric much more robust to translation.

Most uses of deep perceptual similarity and deep perceptual loss have directly compared the corresponding activations of the two images.
This method, referred to as spatial DPS, is formalized as the distance measure between $x$ and $x_0 $ in Eq.~\ref{eq:spatial_metric}, where $f$ is a norm such as L1 or L2 and $p$ is a convolutional feature extractor with extraction layers $l \in L$ each with $C_l$ channels with height $H_l$, and width $W_l$.
%This clearly introduces a dependence on translation as the position in space is taken into account.
%The poor performance of such metrics on translation and rotation motivates the use of spatially invariant methods.
\begin{equation}
    \label{eq:spatial_metric}
    d(x,x_0) = \sum^L_l\frac{1}{C_lH_lW_l} \sum^{C_l,H_l,W_l}_{c,h,w} f(p(x)_{lc}^{hw}-p(x_0)_{lc}^{hw})
\end{equation}

%This work proposes, instead, to calculate deep perceptual similarity by sorting all features in the channel by magnitude and then comparing the feature from one image with the feature from the other image that has the same position in the order.

This work evaluates two additional methods of calculating deep perceptual similarity besides the spatial method.
These two are the mean method tested in~\cite{kumar2022surprising} and a sort method which is introduced in this work.
The two methods are formalized in Eq.~\ref{eq:mean_metric} Eq.~\ref{eq:sorted_metric} where $\overline{x}$ and $x^\downarrow$ are the average and descending reordering of $x$ respectively.
\begin{equation}
    \label{eq:mean_metric}
    d(x,x_0) = \sum^L_l\frac{1}{C_l} \sum^{C_l}_{c} f(\overline{p(x)_{lc}} - \overline{p(x_0)_{lc}})
\end{equation}
\begin{equation}
    \label{eq:sorted_metric}
    d(x,x_0) = \sum^L_l\frac{1}{C_l} \sum^{C_l}_{c} f(p(x)_{lc}^\downarrow-p(x_0)_{lc}^\downarrow)
\end{equation}

Both of these methods ignore the spatial positions of the features.
The mean method compares the average of the features in each channel and the sort method pairs the features of each channel with one another in such a way as to minimize the norm.
In the sort method the norm is minimized for any convex function $f$, compared to any other ordering of the features.
This follows from $x \prec y \rightarrow \sum f(x) \leq \sum f(y)$ and $a^\downarrow - b^\downarrow = a^\downarrow + (-b)^\uparrow \prec a+b$~\cite{marshall2011inequalities}. 

In the case of infinitely large input images and translation-invariant CNNs the two presented methods are translation-invariant as no matter how much the image is translated the same features will appear.
In the case with bounded images, as long as regions with the strongest feature activations aren't shifted off the image or too close to the boundaries, this should still likely result in a metric that is robust to translations.
Even though many CNNs aren't strictly translation-invariant, in general translations have very little effect on the methods.
The reasoning behind comparing average and sorted channels is that a strong activation in one channel often represent different concepts than a similar activation in another.

A problem with the mean and sort methods on their own is that humans would likely say that a lower translation is more similar to the original than a greater one.
As such complete translation-invariance is not desirable.
Thus, this work also investigates metrics that uses the sum of the spatial method with one of the two non-spatial methods.

\subsection{Experimental Setup}
\label{toc:experimental_setup}
DPS relies on neural networks which deep features contain useful information for image comparison.
While networks can be trained specifically for the task, the most common use of DPS and deep perceptual loss is pretrained networks.

This work uses mostly the same feature extraction and comparison setup as~\cite{zhang2018unreasonable}.
The methods are analyzed and evaluated with the L2-norm as the comparison function ($f$) using three models ($p$) pretrained on the ImageNet dataset~\cite{deng2009imagenet}.
The architectures for the three models are SqueezeNet~\cite{iandola2016squeezenet}, AlexNet~\cite{krizhevsky2012imagenet, krizhevsky2014one}, and VGG-16~\cite{simonyan2015very}.
The deep features are extracted from the same multiple layers for each network as in~\cite{zhang2018unreasonable}.
The features extracted in the original work were channel-wise unit-normalized, and this work analyzes and evaluates both using and ignoring this practice.
However, for brevity the analysis in Section~\ref{sec:pcp} concerns only the case without unit-normalization and the use of unit-normalization is later discussed in Section~\ref{toc:analysis}.

%%%%%%%%%%%%%%%%%%%%%%%%%%%%%%%%%%%%%%%%
% DPS pattern analysis section
%%%%%%%%%%%%%%%%%%%%%%%%%%%%%%%%%%%%%%%%

\section{Qualitative Analysis of DPS on Distortions}
\label{sec:pcp}

This work carries out a qualitative analysis of deep perceptual similarity metrics over images specifically designed to test for its strengths and potential flaws.
The analysis is carried out by distorting images in ways for which DPS is previously known to work well or speculated to perform poorly.
The similarity of the distorted image with the original is then compared to the similarities of a set of reference images and the original, where the reference images are intended to be notably less similar than the distortion.
The feature maps at various layers of the DPS networks are then analyzed for each case to gain a deeper understanding of why the metric performed the way it did in this case.
Such insight was then used to create further image pairs to test against.
Finally, for one category of images, specific reference images were created for each image pair.
These reference images, like the others, were created to be perceived by humans as less similar than the distorted versions but intended to fool some DPS metrics.

\begin{figure}[t] % Always use [t] or [!t]
    \includegraphics[width=\columnwidth]{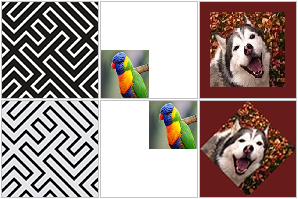}
    \caption{One image (above) and its distorted version (below) from each of the inversion, translation, and rotation categories (left to right).}
    \label{fig:image_pairs}
\end{figure}

The images used in the tests are $96\times96$ pixels and have been designed and distorted by hand.
The distortion tested are divided into four categories; color inversion, translation, rotation, and color stain.
Seven reference images were created; mono-colored images of black, white, gray, red, green, and blue, as well as one with randomly colored pixels.
One image pair each from the inversion, translation, and rotation categories is shown in Fig.~\ref{fig:image_pairs}

\subsection{Black-and-White Color Inversion}
Color inversion of black-and-white images is typically used as an example of when pixel-wise metrics break down.
This is because each pixel in the inverted image is as different from each other as they could be, which means every other possible image would be regarded as more similar, which is obviously not the case for perceptual similarity.
Despite this being used as an example of why to abandon pixel-wise metrics in favor of DPS metrics, there has been little investigation of how well DPS performs in these scenarios.
For these reasons the first set of images created for analyzing DPS were simple black-and-white patterns that were distorted by inverting the colors.

While pixel-wise metrics fail by definition on this category of images, all tested DPS metrics correctly identify each image pair as more similar than any of the reference images.
Analysis of the feature maps reveals that many channels are activated by contrasts or higher-level structures like lines or shapes.
These activations are often completely agnostic to inversion and identify the structures regardless of color.
This makes the black-and-white inversion pairs almost exactly the same for many channels in the feature space, which leads to the good performance of DPS on color inversion.

\subsection{Translation and Rotation}
It is also clear from the feature maps that all activations are strongly spatially correlated to where those features appear in the input image, which can be seen in Fig.~\ref{fig:color_stain_image}.
This is obvious as CNN architectures in general are built around each activation depending only on a small region of the input or previous layer.
While, in theory, activations in the later layers depend on information aggregated from a large swath of the image, in practice, strong activations in the feature maps at any layer are correlated with features in the spatially corresponding region of the input image.
This has been previously suggested as a potential flaw of spatial DPS~\cite{kumar2022surprising}.

To investigate whether this would have a significant impact on spatial DPS and whether other DPS metrics could handle these cases, the categories of translation and rotation have been tested.
The translation images have a region containing much structure in otherwise plain images which have been distorted by translating that region.
The rotation images are simply images that have been distorted by rotation in steps of $22.5$ up to 90 degrees, as well as one rotated 180 degrees.
The purpose of the incremental steps was to see if and further how sensitive DPS is to rotation.

Both the pixel-wise metric and spatial DPS fail to identify any translated image as more similar than the reference images, while the other DPS metrics succeed in each case.
For rotation, both pixel-wise and spatial DPS metrics fail on about the same amount of cases, slightly less than half, while the other DPS metrics almost succeed on each image pair.

This clearly shows that the spatial DPS metric on its own is not suitable for these types of scenarios, while translation-invariant DPS metrics can handle them very well.
It is also interesting that the translation-invariant DPS metrics handle rotation so well since it is well known that early channels in CNNs often learn to identify specific orientations in lines and other structures.
Likely, the later layers of CNNs combine orientation-specific features into higher-level orientation-independent ones.

\subsection{Color Stain}
Another revelation from feature map analysis is that many channels tend to activate strongly from specific colors, textures, or random noisy structures.
This might be challenging for non-spatial methods as ignoring the spatial position of activations might lead to confusing noise for interesting structures.
For example, in the case of mean DPS, an image with a small interesting region might be seen as dissimilar from the same with added stains since the average activation in the noisy one will be larger.

\begin{figure}[t] % Always use [t] or [!t]
    \includegraphics[width=\columnwidth]{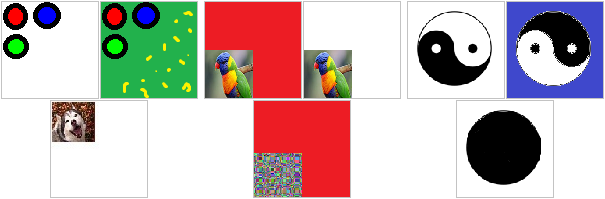}
    \caption{Image pairs from the color stain category (above) with their specific reference images (below).}
    \label{fig:color_noise}
\end{figure}

To test for this the color stain category is used.
The image pair for the color stain category consists of a plain image with a structurally interesting region, and a distorted version with a similar or same interesting region but the plain color is changed, and added noisy features for some images.
The color stain category does not use the same reference images as the other categories, and instead, each image pair has a specific reference image designed to be less perceptually similar.
These reference images have the same plain color without stains as the non-distorted image, but their interesting region is significantly different compared to the distorted version.
Examples of image pairs and their specific reference image are shown in Fig.~\ref{fig:color_noise}.

For the color stain category, the pixel-wise metric again fails for each image pair.
Notably, both the mean and sum of spatial and mean DPS fails almost all image pairs.
The remaining DPS metrics tested perform well, with spatial DPS being the best.

\begin{figure}[t] % Always use [t] or [!t]
    \includegraphics[width=\columnwidth]{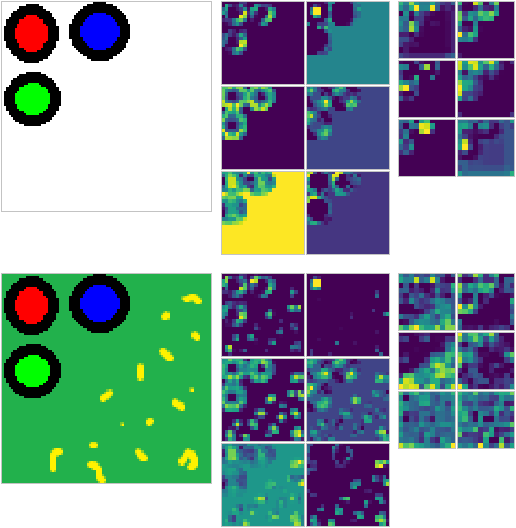}
    \caption{An image (top-left) and its color stain distorted version (bottom-left) with their respective feature maps from the second (middle) and fourth (right) ReLU layer.}
    \label{fig:color_stain_image}
\end{figure}

One specific image in this category was a white image with a red, green, and blue irregular circle in one corner.
The distorted image retained the circles but the plain white background was colored a darker shade of green with random yellow stains.
By observing the feature maps of these images it is clear that the color change and stains add significant activations to the otherwise sparse feature maps, especially in later layers.
This is shown in Fig.~\ref{fig:color_stain_image}, where the image and its distortion are displayed together with feature maps from the second and fourth SqueezeNet ReLU layer.

\section{Evaluation}
In order to investigate how the insights gained through the qualitative analysis translate to performance on a perceptual similarity dataset, the DPS metrics are evaluated on the BAPPS dataset, using the same procedure as in the original work~\cite{zhang2018unreasonable}.
However, the evaluation is only carried out for the pretrained networks SqueezeNet, AlexNet, and VGG-16 without additional fitting to the training data.
This is equivalent to what the original work refers to as "Net (Supervised)", with the addition of testing non-spatial DPS metrics.
This evaluation follows the original work which means $f$ is the L2 norm and the features extracted from $p$ have been channel-wise unit-normalized.

\subsection{BAPPS} %Currently from other paper
BAPPS is an image dataset consisting of $64\times64$ image patches sampled from the MIT-Adobe 5k~\cite{bychkovsky2011learning}, RAISE1k~\cite{dang2015raise}, DIV2K~\cite{agustsson2017ntire}, Davis Middleburry~\cite{scharstein2001taxonomy}, video deblurring~\cite{su2017deep}, and ImageNet~\cite{deng2009imagenet} datasets as well as a host of distortions of those same patches.
The BAPPS dataset consists of two sets with different labels and intended use, Two Alternative Forced Choice (2AFC) and Just Noticeable Differences (JND).

2AFC consists of image patches and two distorted versions of each patch, as well as human annotations as to which distorted patch is most similar to the original.
The aim of 2AFC is to train and evaluate models for perceptual similarity judgment by evaluating if those models give higher similarity to the distortion that most human annotators agreed was more similar.
In addition to evaluating on the complete 2AFC part, results are also given for a number of subdivisions defined by the type of distortions that are applied: (1) \textbf{Traditional} augmentation methods, outputs from (2) \textbf{CNN-based} autoencoders, (3) \textbf{superresolution}, (4) \textbf{frame interpolation}, (5) \textbf{video deblurring}, and (6) \textbf{colorization}.

JND consists of an image patch as well as a barely distorted version along with human annotations of whether the two patches are the same.
The human annotators were shown the two images only briefly and were also shown pairs of the same and very different images.
The aim of JND is to test models for perceptual similarity by evaluating if those models give a higher similarity to those samples that human annotators had difficulty telling apart.

% Results
\section{Results}
\label{toc:results}

An aggregation of the outcome of the tests described in Section~\ref{sec:pcp} is shown in Table~\ref{tab:analysis_results}.
The performance is presented as the number of images, where the metric did not find any of the reference images to be more similar than the distorted version.

The results of the evaluation on the BAPPS dataset are shown in Table~\ref{tab:bapps_results} for each subdivision of the 2AFC part, for the entire 2AFC part given by the average over the subdivisions, and for the JND set.
The results for the evaluated metrics are presented along with the LPIPS metrics from~\cite{zhang2018unreasonable} and human performance for reference.

\begin{table}[t]
    \caption{Results on analyzed image pairs for different metrics}
    \label{tab:analysis_results}
    \begin{center}
    \begin{small}
    \begin{tabular}{l l c c c c }
    \toprule
        %Method & Network & Trad- & CNN-& All & Super- & Video & Color-& Frame & All & All \\
        % & & itional & based & & res & Deblur & ization & Interp & & \\
        Method & Network & \makecell{Inv-\\ert} & \makecell{Rot-\\ate} & \makecell{Tran-\\slate} &\makecell{Color\\Stain}\\\midrule
        Pixel-Wise & \multicolumn{1}{c}{-} & 0/15 & 17/30 & 0/5 & 0/5 \\\midrule
         & SqueezeNet           & 11/11 & 20/30 & 0/5 & 5/5 \\
        Spatial & AlexNet       & 11/11 & 11/30 & 0/5 & 3/5 \\
         & VGG-16               & 10/11 &  6/30 & 0/5 & 4/5\\\midrule
         & SqueezeNet           & 11/11 & 28/30 & 5/5 & 4/5\\
        Sort & AlexNet          & 11/11 & 30/30 & 5/5 & 2/5\\
         & VGG-16               & 11/11 & 30/30 & 5/5 & 3/5 \\\midrule
         & SqueezeNet           & 11/11 & 29/30 & 5/5 & 3/5\\
        Mean & AlexNet          & 11/11 & 28/30 & 5/5 & 2/5\\
         & VGG-16               & 10/11 & 30/30 & 5/5 & 1/5\\\midrule
         & SqueezeNet           & 11/11 & 22/30 & 0/5 & 5/5\\
        Spatial+Sort & AlexNet  & 11/11 & 19/30 & 0/5 & 4/5\\
         & VGG-16               & 11/11 & 21/30 & 1/5 & 4/5 \\\midrule
         & SqueezeNet           & 11/11 & 22/30 & 0/5 & 5/5\\
        Spatial+Mean & AlexNet  & 11/11 & 15/30 & 0/5 & 2/5\\
         & VGG-16               & 10/11 &  9/30 & 0/5 & 4/5\\
         \bottomrule

    \end{tabular}
    \end{small}
    \end{center}
\end{table}

%\begin{table}[t]
%    \caption{Results on analyzed image pairs for different metrics}
%    \label{tab:analysis_results}
%    \begin{center}
%    \begin{small}
%    %\begin{sc}
%    \begin{tabular}{l l c c c c }
%    \toprule
%        %Method & Network & Trad- & CNN-& All & Super- & Video & Color-& Frame & All & All \\
%        % & & itional & based & & res & Deblur & ization & Interp & & \\
%        Method & Network & \makecell{Inv-\\ert} & \makecell{Rot-\\ate} & \makecell{Tran-\\slate} &\makecell{Color\\Noise}\\\midrule
%        Pixel-Wise & \multicolumn{1}{c}{-} & 0/15 & 17/30 & 0/5 & 0/5 \\\midrule
%         & SqueezeNet & 15/15 & 22/30 & 0/5 & 5/5 \\
%        Spatial & AlexNet & 15/15 & {19/30} & 0/5 & 4/5 \\
%         & VGG-16 & 15/15 & 15/30 & 0/5 & 4/5\\\midrule
%         & SqueezeNet & 15/15 & 29/30 & 5/5 & 4/5\\
%        Sort & AlexNet & 15/15 & 27/30 & 5/5 & 3/5\\
%         & VGG-16 & 15/15 & 26/30 & 5/5 & 4/5 \\\midrule
%         & SqueezeNet & 15/15 & 27/30 & 5/5 & 0/5\\
%        Mean & AlexNet & 15/15 & 26/30 & 5/5 & 2/5\\
%         & VGG-16 & 15/15 & 30/30 & 5/5 & 0/5\\
%         \midrule
%         & SqueezeNet & 15/15 & 29/30 & 5/5 & 4/5\\
%        Spatial+Sort & AlexNet & 15/15 & 27/30 & 5/5 & 3/5\\
%         & VGG-16 & 15/15 & 26/30 & 5/5 & 4/5 \\\midrule
%         & SqueezeNet & 15/15 & 27/30 & 5/5  & 0/5\\
%        Spatial+Mean & AlexNet & 15/15 & 26/30 & 5/5 & 2/5\\
%         & VGG-16 & 15/15 & 30/30 & 5/5 & 0/5\\
%         \bottomrule
%
%    \end{tabular}
%    %\end{sc}
%    \end{small}
%    \end{center}
%\end{table}

\begin{table*}[t] % TODO replace with t
    \caption{Results on the BAPPS validation set}
    \label{tab:bapps_results}
    \begin{center}
    \begin{small}
    %\begin{sc}
    \begin{tabular}{l l c c c c c c c c c c}
    \toprule
         & & \multicolumn{3}{c}{Distortions} & \multicolumn{5}{c}{Real Algorithms} & All & JND\\ \cmidrule(lr){3-5}\cmidrule(lr){6-10}\cmidrule(lr){11-11}\cmidrule(lr){12-12}
        %Method & Network & Trad- & CNN-& All & Super- & Video & Color-& Frame & All & All \\
        % & & itional & based & & res & Deblur & ization & Interp & & \\
        Method & Network & \makecell{Trad-\\itional} & \makecell{CNN-\\based} & All & \makecell{Super-\\res} & \makecell{Video\\Deblur} & \makecell{Color-\\ization} & \makecell{Frame\\Interp} & All & All & JND\\\midrule
        Human & - & 80.8 & 84.4 & 82.6 & 73.4 & 67.1 & 68.8 & 68.6 & 69.5 & 73.9 & -\\\midrule
         & SqueezeNet & \textcolor{gray}{76.1} & \textcolor{gray}{83.5} & \textcolor{gray}{79.8} & 71.1 & 60.8 & 65.3 & 63.2 & 65.1 & \textcolor{gray}{70.0} & -\\
        LPIPS*~\cite{zhang2018unreasonable} & AlexNet & \textcolor{gray}{77.6} & \textcolor{gray}{82.8} & \textcolor{gray}{80.2} & 71.1 & \textbf{61.0} & \textbf{65.6} & \textbf{63.3} & 65.2 & \textcolor{gray}{70.2} & -\\
         & VGG-16 & \textcolor{gray}{77.9} & \textcolor{gray}{83.7} & \textcolor{gray}{80.8} & 71.1 & 60.6 & 64.0 & 62.9 & 64.6 & \textcolor{gray}{70.0} & -\\\midrule
         & SqueezeNet & 73.3 & 82.6 & 78.0 & 70.1 & 60.1 & 63.6 & 62.0 & 64.0 & 68.6 & 60.2\\
        Spatial & AlexNet & 70.6 & \textbf{83.1} & 76.8 & 71.7 & 60.7 & 65.0 & 62.7 & 65.0 & 68.9 & 57.6\\
         & VGG-16 & 70.1 & 81.3 & 75.7 & 69.0 & 59.0 & 60.2 & 62.1 & 62.6 & 67.0 & 59.1\\\midrule
        
          & SqueezeNet & 77.1 & 82.3 & 79.7 & 69.9 & 60.0 & 65.2 & 63.1 & 64.5 & 69.5 & 63.6\\
        Mean & AlexNet & 73.9 & 82.8 & 78.4 & 71.4 & 60.7 & 65.5 & 63.5 & \textbf{65.3} & \textbf{69.6} & 60.2\\
         & VGG-16 & 77.9 & 81.8 & \textbf{79.8} & 68.9 & 59.5 & 64.0 & 63.0 & 63.8 & 69.2 & \textbf{65.2}\\\midrule
          & SqueezeNet & 76.8 & 82.0 & 79.4 & 69.8 & 60.1 & 64.6 & 61.9 & 64.1 & 69.2 & 62.0\\
        Sort & AlexNet & 73.3 & 82.8 & 78.0 & 71.1 & 60.6 & 64.6 & 62.6 & 64.7 & 69.2 & 58.5\\
         & VGG-16 & \textbf{78.1} & 81.5 & \textbf{79.8} & 68.1 & 59.2 & 62.7 & 61.5 & 62.9 & 68.5 & 64.8\\\midrule
         & SqueezeNet & 75.0 & 82.5 & 78.8 & 69.9 & 60.1 & 64.5 & 62.1 & 64.2 & 69.0 & 61.5\\
        Spatial+Mean & AlexNet & 71.8 & 83.0 & 77.4 & 71.6 & 60.7 & 65.5 & 62.7 & 65.1 & 69.2 & 58.5\\
         & VGG-16 & 73.4 & 81.9 & 77.7 & 69.3 & 59.4 & 64.5 & 62.5 & 63.9 & 68.2 & 61.0\\\midrule
         & SqueezeNet & 75.5 & 82.5 & 79.0 & 70.0 & 60.1 & 64.4 & 61.9 & 64.1 & 69.1 & 61.2\\
        Spatial+Sort & AlexNet & 72.2 & \textbf{83.1} & 77.7 & 71.3 & 60.6 & 64.9 & 62.8 & 64.9 & \textbf{69.2} & 58.5\\
         & VGG-16 & 74.9 & 81.9 & 78.4 & 69.4 & 59.4 & 62.3 & 62.1 & 63.3 & 68.4 & 61.9\\\bottomrule
        \multicolumn{12}{l}{\makecell[l]{*LPIPS networks have been trained for image similarity on traditional and CNN-based distortions while the other\\models have not been trained for image similarity at all, this gives them a significant advantage when testing on\\the same distortion types. To indicate this, such values have been grayed out. The LPIPS rows presented only\\considers the best overall LPIPS row in the original work. Additionally, the LPIPS results are taken from the\\original work whereas all other results have been collected from new experiments.}}
    \end{tabular}
    %\end{sc}
    \end{small}
    \end{center}
\end{table*}

%For tables see \footnote{\url{https://www.inf.ethz.ch/personal/markusp/teaching/guides/guide-tables.pdf} and \url{https://www.darkhorseanalytics.com/blog/clear-off-the-table apply religiously}}.
%As a general note, it is good practice to make a line break after each period in \LaTeX. 
%This makes it easier to read and modify, and especially to spot sentence construction flaws.

% Here is where you put evidence for all the claims you've made in the abstract - introduction - conclusion parts.
% All plots, tables and whatever else visualizations you have go here and gets to be discussed.
% Try to avoid marketing words such as VERY, EXTREMELY and so on.
% Do no hesitate to break down the section in multiple sub section to give some structure to the presentation of the results!

% Discussion and Future Work
\section{Discussion}
\label{toc:analysis}
% Here we want to write about some important aspects of your observations. These are differing, depending on the paper, e.g.,
% Some typical good cases and failure cases; or cases where the methods disagree
% Looking into the actual activations of several layers, what is important for the method to make it work. Is there any unwanted bias towards the data
% Research is not about performance, but about hypotheses evaluation, investigation, and understanding. 
% What does the results and analysis lead us to believe, but that we cannot yet prove.
% How do we think the field needs to develop?
% Other interesting things that can be discussed that aren't necessarily analysis of the results

%% Discussion on analysis

% Spatial DPS by itself clearly insufficient.
% Which cases did each metric fail on, which did only a few fail on
% Why did sort handle cases that mean couldn't
% The combined DPS doesn't seem to get the best of both worlds
The purpose of this work has been to evaluate if and how DPS metrics can handle the typical cases where pixel-wise metrics fail, and to investigate whether similar flaws exist in current DPS implementations.
All tested DPS metrics handle the color inversion tests for which pixel-wise metrics break down, and additionally the clear preference for contrasts and structures in feature maps indicates that DPS metrics are well-suited to handle other similar color-changing operations.
By far most common form of DPS metrics used is spatial DPS, which seems to perform as poorly as the pixel-wise metric on the rotation and translation test cases.
While the non-spatial DPS metrics perform well on these weaknesses, they do not perform as well as spatial metrics on the color stain category of tests.
This is especially true for mean DPS which failed most of the color stain tests.
The spatial and non-spatial combined metrics perform similar to spatial DPS, indicating that perhaps combining metrics using unweighted summation gives a preference for spatial DPS.
Though the results of the combined metrics improved somewhat for rotation, indicating that there are at least some benefits to this strategy.

%%Discussion on analysis+BAPPS
%  Mean problems not represented by BAPPS trial
%  Perhaps BAPPS doesn't cover scenarios where that comes into play
%  Perhaps such scenarios are not common in reeal datsets
%  Perhaps plain images that most of our tests were are not so common in real datasets
% Selecting the right DPS metric more influential than networks... in this case. However other research, cite google (and the LNA paper when that comes to arxiv), show that actually the architecture and layer selection has even stronger effect

Analyzing the BAPPS scores for the different DPS metrics shows that flaws of spatial DPS also affect performance on a perceptual similarity dataset.
While spatial DPS, in general, performs worse than the other DPS metrics, notably, this is especially true for the traditional augmentations subdivision.
Traditional augmentations include operations such as rotation, translation, and skewing which indicates that the weaker performance of spatial DPS is due to the flaws identified in this work.

Another notable result is that mean DPS, on average, performs best on BAPPS, even though it was vulnerable to the color stain category of distortions to a much larger degree than sort DPS.
However, both mean and sort DPS metrics perform similarly and are both better choices than spatial DPS.
It is possible that color stain and related distortions are not so common to be a problem in a real-world scenario, or that the BAPPS dataset does not include such cases.
Additionally, the image pairs used for analysis in this work have often been very simple and plain compared to the images typically included in datasets.

\subsection{The effects of unit-normalization}
As mentioned in Subsection~\ref{toc:experimental_setup}, the qualitative analysis described in Section~\ref{sec:pcp} was also performed with channel-wise unit-normalization of the extracted features.
This had three notable effects.
First, the success rate in the rotate category rose for all DPS metrics, especially for the metrics that use spatial DPS.
In fact, those metrics became almost competitive with mean and sort DPS.
Second, while spatial still fails on each image pair in the translate category, the combined metrics are somewhat improved.
Likely due to normalizing making the spatial distances lower which gives more weight to the non-spatial distances.
Finally, using normalization made each DPS metric perform poorly in the color stain category.

When evaluating with the BAPPS dataset unit-normalization has only a small positive effect on performance.
Likely translation and rotation and similar augmentation are more common in that dataset than augmentations similar to the color stain procedures.

%% Short paper, maybe skip conclusion
% So what? I've read your paper and now what is the take away message?
% Here you want to have a couple of sentences makes it to the point without additional bla-bla.
% -> keep it short!

\section{Future Work}
\label{toc:future}
%% Future Work
% Explore use of non-spatial DPS in the use cases where spatial DPS is currently SOTA
% Examine why the color noise problem does not affect BAPPS performance (?)
% Why doesn't complete translation invariance hurt results on BAPPS
% Investigate how to best combine spatial and global DPS metrics
% Use understanding gained to develop a new DPS metric that integrates both spatial and global information without hacking the two together (?)

From the results and analysis presented in this work there are some notable directions of research to explore.

Both this and a prior work~\cite{kumar2022surprising} has shown that spatial DPS does not perform as well as on perceptual similarity tasks as other implementations of DPS.
One future possibility is to investigate if this translates to related field such as deep perceptual loss and content-based image retrieval.
If it does, simply changing the way perceptual loss is calculated could improve the results on many different tasks.

While most DPS metrics outperform previous perceptual similarity metrics, the discrepancy in performance of DPS metrics indicates that exploring how to calculate DPS metrics is an open problem.
For example, a DPS metric that make use of both spatial and non-spatial comparisons could perhaps gain the benefit of both.
Additionally, the upsides and downsides of unit-normalization remain inconclusive.

\bibliography{biblio}
\bibliographystyle{IEEEtranN}

\end{document}